\documentclass{article} 
\usepackage{iclr2027_conference,times}

\usepackage{amsmath,amsfonts,bm}

\def\eqref#1{equation~\ref{#1}}

\def\1{\bm{1}}

\DeclareMathAlphabet{\mathsfit}{\encodingdefault}{\sfdefault}{m}{sl}
\SetMathAlphabet{\mathsfit}{bold}{\encodingdefault}{\sfdefault}{bx}{n}

\usepackage{hyperref}
\usepackage{url}
\usepackage{amsmath}
\usepackage{amssymb}
\usepackage{booktabs}
\usepackage{multirow}
\usepackage{graphicx}
\usepackage{enumitem}
\usepackage{booktabs}
\usepackage{xcolor}
\usepackage{colortbl}
\definecolor{oursrow}{HTML}{F1ECF9} 
\usepackage{wrapfig}

\title{EASE: Behavior-Adaptive Skill Curation for Self-Evolving Agents}

\author{Zhen Xiong$^{1}$, Qiaoyu Tan$^{2}$ \\
$^{1}$ New York University, $^{2}$ New York University Shanghai}

\iclrfinalcopy 
\begin{document}

\maketitle

\begin{abstract}
Agent skills provide a lightweight mechanism for self-evolving agents to accumulate reusable procedural knowledge without updating model parameters. However, existing learned skill curators typically optimize curation without explicitly modeling the behavior of the downstream executor. We show that this limitation can lead to systematic cross-executor degradation: curators optimized with different executors perform best when paired with their own training executor, suggesting that effective skill curation is executor-dependent. To address this challenge, we formulate behavior-adaptive skill curation and introduce \textbf{EASE}, a framework that learns a single curator capable of adapting its curation decisions to different executor behaviors. EASE maintains an online behavioral profile summarizing recent execution patterns and conditions the curator on this profile, together with the current trajectory and retrieved skills, to dynamically add, modify, or remove skills from an evolving repository. We train the shared curator jointly across multiple frozen executors with reinforcement learning, using retrieval-aware and behavior-aware temporal attribution to focus optimization on curation actions with observable downstream influence. Across multiple public agentic benchmarks, including ALFWorld, ScienceWorld, and WebShop, and heterogeneous executor families ranging from Qwen3-8B/32B and GPT-OSS-120B to previously unseen Kimi K2.6, DeepSeek V4 Flash, and Gemini 3.5 Flash, EASE outperforms strong skill- and memory-based baselines without per-executor finetuning. Beyond task performance, behavior-adaptive curation also enables more efficient skill evolution: across the three benchmarks, EASE maintains $34.5{\sim}41.0$\% fewer skills, improves skill retrieval by $36.3{\sim}38.7$\% and measured edit utility by $51.8{\sim}60.0$\%, and reduces deployment-time inference tokens by $9.1{\sim}14.5$\%. These results establish behavior-adaptive skill curation as an effective principle for building self-evolving agents. Code is available at \url{https://github.com/Eric2i/EASE}.

\end{abstract}

\section{Introduction}
Large language model (LLM) agents increasingly operate in interactive environments that require long-horizon reasoning, tool use, and continual adaptation from experience \citep{yang2024sweagent,hu2025osagents}. While model-based finetuning can improve an agent over time, repeatedly updating model parameters during deployment is often impractical. The executor may be accessible only through an API, optimization can be computationally expensive, and continual weight updates may introduce instability or overfitting. A lightweight alternative is to accumulate experience as agent skills, namely compact and reusable descriptions of procedural knowledge that can be retrieved and reused in future tasks without modifying the executor itself \citep{anthropic2025agentskills,ma2026skillgen}. Recent work has therefore explored automatically constructing and curating evolving skill repositories from agent trajectories and interaction feedback \citep{xia2026skillrl,ni2026trace2skill,wang2026skillx,wang2026sage,ouyang2026skillos}.


A key challenge, however, is that effective skill curation may not be independent of the executor that consumes those skills. Different executors can exhibit different action preferences, failure modes, recovery behaviors, reasoning patterns, and abilities to exploit retrieved instructions. As a result, the same skill or curation decision may not have the same downstream utility across executors. Existing learned skill curators \citep{wang2026sage,ouyang2026skillos} typically optimize curation without explicitly modeling such executor-specific behavior, which can make their decisions brittle when the deployment executor changes.

\begin{wrapfigure}{r}{0.5\textwidth}
    \centering
    \vspace{-8pt}
    \includegraphics[width=\linewidth]{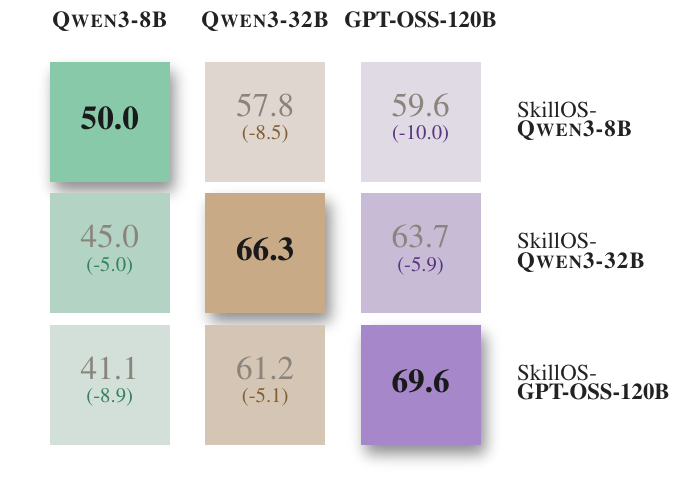}
    \caption{\textbf{Skill curation is executor-dependent.} We independently train a skill curator with each executor and evaluate all curator--executor pairings. Matched pairs consistently outperform mismatched ones, suggesting that effective skill curation depends on downstream executor behavior and motivating behavior-adaptive skill curation.}
    \label{fig:heatmap}
    \vspace{-12pt}
\end{wrapfigure}
We observe this phenomenon empirically in Figure~\ref{fig:heatmap}. When separately trained skill curators are paired with different executors, each curator performs best with the executor on which it was optimized, while mismatched curator and executor pairs consistently degrade. This pattern suggests that effective skill curation is executor-dependent. The curator should not only reason about what experience has been collected, but also about how the downstream executor is likely to use that experience.

This motivates a distinct problem that we call \emph{behavior-adaptive skill curation}: learning a shared curator that dynamically adjusts its curation decisions according to the observed behavior of the executor that will consume the resulting skills. A straightforward solution would be to train a dedicated curator for each executor, but this is costly and cannot support newly released or otherwise unseen executors. Pooling data from multiple executors is also insufficient by itself, because the same curation decision may receive conflicting supervision under different executor behaviors. Conditioning on model identity or static metadata may help for known executors, but provides limited information about how an unseen executor actually behaves during interaction. We therefore argue that adaptation should be driven by behavioral evidence inferred online from execution itself.

To address this challenge, we introduce EASE, a framework for behavior-adaptive skill curation in self-evolving agents. EASE continuously summarizes an executor's recent interaction patterns into an online behavioral profile. A shared curator conditions on this profile, together with the current trajectory and retrieved skills, to dynamically add, modify, or remove skills from the evolving repository. We train the curator jointly across multiple frozen executors with reinforcement learning, encouraging it to learn how curation decisions should vary with executor behavior rather than overfitting to any single executor. EASE further uses retrieval- and behavior-aware temporal credit assignment to focus optimization on curation actions with observable downstream influence. At deployment, both curator and executor parameters remain fixed, while adaptation proceeds through updates to the behavioral profile and skill repository. We evaluate EASE across multiple interactive agent benchmarks and heterogeneous executor families, including both executors seen during curator training and previously unseen executors. A single EASE curator matches or improves executor-specific curators on seen executors and generalizes to unseen ones without per-executor finetuning. Beyond task performance, behavior-adaptive curation also produces more compact and useful skill repositories, with higher skill retrieval and edit utility as well as lower deployment-time inference overhead. These results suggest that explicitly adapting skill curation to executor behavior provides a more robust foundation for self-evolving agents whose underlying executors may change over time. Our main \textbf{contributions} are summarized as follows:
\begin{itemize}[leftmargin=*, topsep=0mm]
    \item \textbf{Behavior-adaptive skill curation.} We identify a fundamental challenge in self-evolving skill curation: effective curation is not executor-independent, but changes with the behavior of the downstream executor. We formulate behavior-adaptive skill curation as a new paradigm in which skill evolution explicitly conditions its curation decisions on observed executor behavior, rather than assuming that a fixed curation policy transfers uniformly across executors.
    \item \textbf{EASE.} We introduce a shared behavior-adaptive skill curator that continuously summarizes recent execution patterns into an online behavioral profile and uses this profile to guide skill addition, modification, and deletion. EASE is trained jointly across multiple frozen executors with reinforcement learning, together with retrieval- and behavior-aware temporal credit assignment that focuses optimization on curation actions with observable downstream influence.
    \item \textbf{Broad empirical advantage.} We show that EASE generalizes across heterogeneous executors and multiple interactive environments, including executors unseen during curator training, without per-executor finetuning. Beyond task performance, EASE produces more compact and useful skill repositories, with higher skill retrieval and edit utility and lower deployment-time inference overhead, demonstrating that behavior adaptation improves both the effectiveness and efficiency of self-evolving skill curation.
\end{itemize}

\section{Related Work}
\paragraph{Automated agent skills.}
Prior work enables agents to reuse past experience through reflections, reusable insights, executable skills, workflows, and procedural memories
\citep{shinn2023reflexion,zhao2024expel,wang2023voyager,wang2025awm,zhang2026ace,ouyang2026reasoningbank,fang2026memp}.
Building on this direction, recent methods increasingly automate the construction and refinement of reusable skills from execution trajectories and feedback
\citep{ni2026trace2skill,wang2026skillx,alzubi2026evoskill,ma2026skillgen,zhang2026coevoskills}.
Several approaches further study whether generated skills are actually useful for downstream execution, for example by evaluating their marginal contribution or repairing low-utility skills
\citep{ma2026skillgen,wang2026assay}.
A complementary line of work moves from optimizing individual skill artifacts to learning policies for skill generation and curation, often through reinforcement learning over sequential interactions
\citep{wang2026sage,ouyang2026skillos,zhang2026sapo,xia2026skillrl}.
EASE builds on this latter direction, but focuses on a different question: how should a learned curator adapt its decisions when the executor that consumes the resulting skills exhibits different execution behavior? To support such adaptation, EASE conditions curation on an online behavioral profile and uses temporal credit assignment to focus learning on curation actions with observable downstream influence.

\paragraph{Skill transferability and executor adaptation.}
Recent work also investigates whether learned skills remain useful when transferred across agents or model backbones.
Trace2Skill, SkillGen, and CoEvoSkills evaluate generated skills on models different from those used to construct them
\citep{ni2026trace2skill,ma2026skillgen,zhang2026coevoskills}, showing that reusable skill artifacts can transfer across executors, while also revealing that their effectiveness may remain model-dependent.
At the policy level, SkillOS studies whether a learned skill curator can transfer across frozen executor backbones
\citep{ouyang2026skillos}.
These studies establish cross-executor transfer as an important capability, but transferability alone does not explicitly model how curation should change with the behavior of the deployment executor.
EASE instead formulates behavior-adaptive skill curation: a single shared curator observes an executor's online behavioral profile and adjusts its add, modify, and delete decisions accordingly. This formulation allows adaptation to both seen and previously unseen executors without executor-specific fine-tuning, treating cross-executor generalization as a consequence of behavior-conditioned curation rather than the primary objective.

\section{Method}
\label{sec:method}
\begin{figure}[t]
\centering
\includegraphics[width=\linewidth]{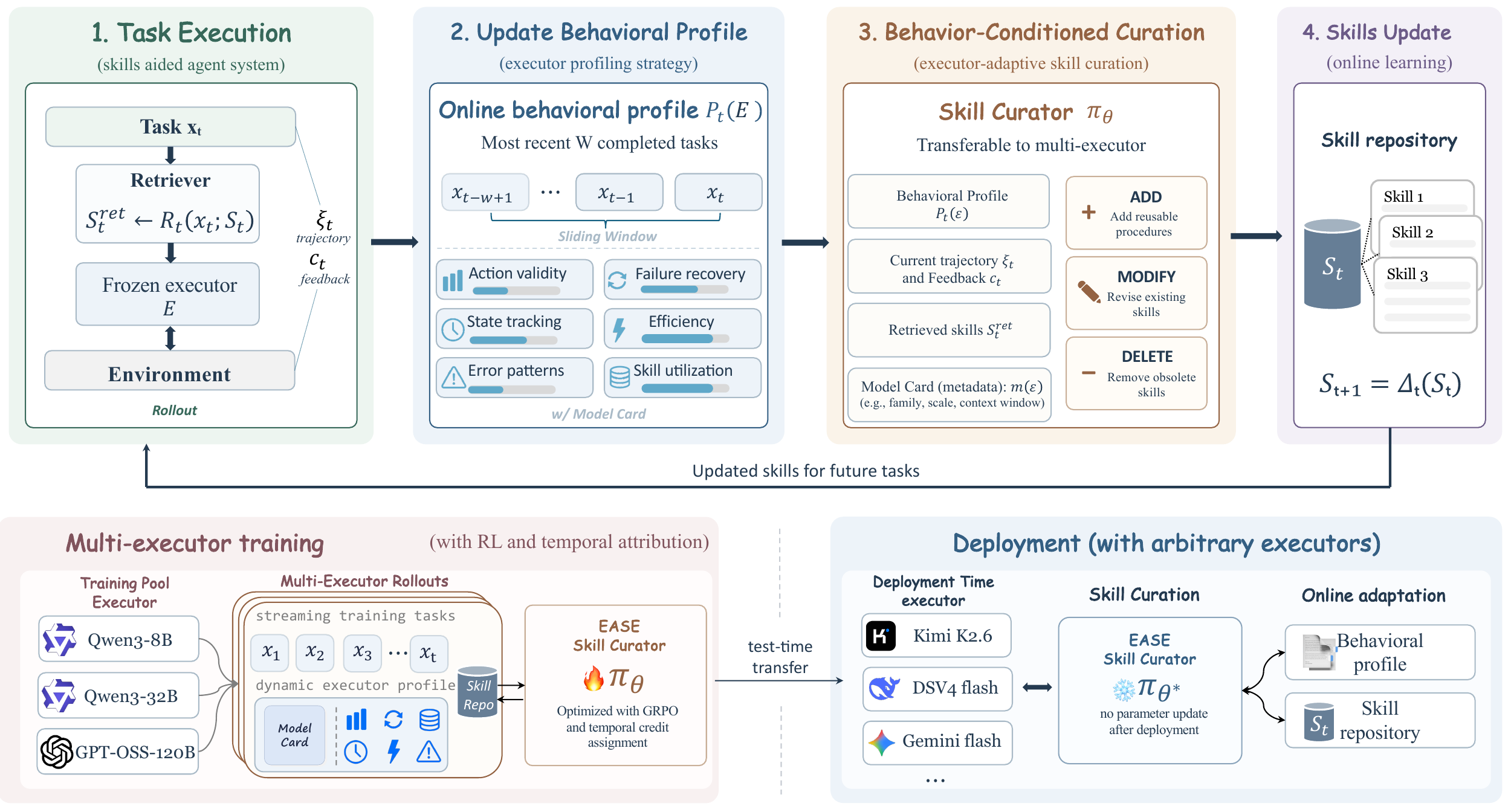}
\caption{\textbf{Overview of EASE framework.}
\emph{Top:} the online loop for one executor. (1) A frozen executor $E$ solves task $x_t$ with
skills retrieved from the repository $\mathcal{S}_t$. (2) The resulting trajectory and feedback
update the online behavioral profile $P_t(E)$, which summarizes execution statistics over recent $W$ tasks. (3) Conditioned on the executor descriptor $e_t=[P_t(E);m(E)]$, the current
trajectory, and the retrieved skills, a skill curator $\pi_\theta$ performs different operations. (4) The updated repository
$\mathcal{S}_{t+1}$ serves subsequent tasks. \emph{Bottom left:} $\pi_\theta$ is trained with GRPO
across a pool of frozen executors with temporal-attribution gates. \emph{Bottom right:} at deployment,
the frozen curator adapts to unseen executors without per-executor finetuning.}
\label{fig:framework}
\end{figure}

We present EASE, a framework for behavior-adaptive skill curation in self-evolving agents (Figure~\ref{fig:framework}). Instead of training a separate curator for each executor, EASE adapts skill curation to the observed behavior of the executor that consumes the resulting skills.

\subsection{Behavior-Adaptive Skill Curation and Framework Overview}
\label{sec:prelim}

We consider a self-evolving agent composed of a frozen LLM-based executor $E$, an evolving skill repository $\mathcal{S}_t$, and a shared skill curator $\pi_\theta$. Following \citet{ouyang2026skillos}, tasks $x_1,x_2,\ldots$ arrive sequentially, and each task is completed before the next one is observed. Before task $x_t$, the executor retrieves at most $k$ relevant skills $\mathcal{S}^{\mathrm{ret}}_t=\mathcal{R}_k(x_t;\mathcal{S}_t)$. The executor then interacts with the environment, producing an agentic trajectory $\xi_t$ and, when available, feedback $c_t$ from the environment or its own self-assessment. After the task, the curator emits curation operations $\Delta_t$ that update the repository to $\mathcal{S}_{t+1}=\Delta_t(\mathcal{S}_t)$, which is available only to subsequent tasks.

\paragraph{Behavior-adaptive curation.}
Existing learned curators decide $\Delta_t$ from the current experience and retrieved skills alone, $\pi_\theta(\cdot\mid\xi_t,c_t,\mathcal{S}^{\mathrm{ret}}_t)$, without representing how the downstream executor consumes the resulting skills. Since curator effectiveness varies systematically across executors (Figure~\ref{fig:heatmap}), we formulate behavior-adaptive skill curation, in which the curator additionally conditions on a behavioral representation of the executor:
\begin{equation}
\Delta_t\sim\pi_\theta\!\left(\cdot\mid\xi_t,c_t,\mathcal{S}^{\mathrm{ret}}_t,b_t(E)\right),
\label{eq:adaptive-curator}
\end{equation}
where $b_t(E)$ summarizes the executor's behavior observed up to task $t$. Rather than identifying the executor and selecting an executor-specific curator, EASE learns a single conditional policy whose decisions vary with observable behavior, which also applies to previously unseen executors. We instantiate $b_t(E)$ with an online behavioral profile $P_t(E)$, described next.

\subsection{Online Executor Behavioral Profiling and Skill Curation}
\label{sec:conditioning}
\label{sec:profile}

\paragraph{Accumulating behavioral evidence.}
A single trajectory describes one execution, whereas behavior-adaptive curation requires evidence about recurring execution patterns. For each executor and task stream, let $P_t(E)$ denote the behavioral profile after completing task $x_t$. We aggregate evidence over a sliding window containing the most recent $W$ completed tasks:
\begin{equation}
P_t(E)=\operatorname{Agg}\!\left(
\left\{\phi(\xi_i,c_i,\mathcal{S}^{\mathrm{ret}}_i)\right\}_{i=\max(1,t-W+1)}^{t}
\right),\qquad P_0(E)=\varnothing.
\label{eq:profile-update}
\end{equation}
Here, $\phi$ extracts observable event counts and measurements, while $\operatorname{Agg}$ combines them into counts, rates, and averages together with their sample sizes. Before $W$ tasks have been completed, the profile uses all available episodes. The finite window allows the profile to track the executor's recent behavior as the skill repository evolves, rather than treating behavior as a fixed intrinsic property of the underlying model.

The feedback $c_i$ is the executor's self-assessment or feedback exposed by the environment and does not require access to held-out answers. The profile covers action validity, failure recovery, state tracking, execution efficiency, observable error patterns, and skill utilization. Features are computed from logged interactions without additional model calls. Missing signals are treated as unavailable rather than recorded as zero. In particular, retrieval is directly observable, whereas implicit reliance on a skill may not always be identifiable. We therefore distinguish retrieval from explicit usage evidence. Appendix~\ref{app:profile} provides the detailed feature families and aggregation rules.

\paragraph{Behavior-conditioned curation.}
After the current trajectory is incorporated into $P_t(E)$, the curator samples a sequence of curation operations:
\begin{equation}
\Delta_t\sim\pi_\theta\!\left(\cdot\mid\xi_t,c_t,\mathcal{S}^{\mathrm{ret}}_t,e_t\right),
\qquad e_t=[P_t(E);m(E)],
\qquad \mathcal{S}_{t+1}=\Delta_t(\mathcal{S}_t),
\label{eq:curator}
\end{equation}
where $m(E)$ denotes optional static executor metadata.

The curator supports three repository operations: \texttt{add}, \texttt{modify}, and \texttt{delete}, corresponding to inserting a new reusable skill, revising an existing skill, and removing a redundant or counterproductive skill, respectively. The online behavioral profile provides the primary adaptation signal. For example, repeated action rejections may motivate skills with explicit precondition checks, whereas an executor that already handles such constraints reliably may benefit from shorter or higher-level procedural guidance.

When available, static metadata $m(E)$, such as model family, scale, or context window, provides supplementary context. However, it does not replace behavioral evidence and unavailable fields are explicitly marked as unknown. This distinction is important because online behavioral profiles can be constructed for both seen and previously unseen executors from their actual interactions.

At cold start, $P_0(E)$ and the repository are empty. The first curation decision therefore uses the profile formed after the first completed task. Profiles and repositories are maintained independently across executors and rollouts and are never shared between independent streams.

\subsection{Learning a Behavior-Adaptive Curator}
\label{sec:pool}

A behavior-conditioned curator is useful only if it learns how different behavioral patterns should induce different curation decisions. We therefore train only the curator, using experience collected from a pool of frozen executors. For each training group, we sample an executor $E$ and an ordered stream $G=(x_1,\ldots,x_T)$ of $T\geq2$ related tasks, and collect $N$ independent rollouts with the same executor and task sequence, each maintaining its own skill repository, behavioral profile, and execution history. Related tasks allow skills acquired earlier in the stream to influence later tasks, while the streaming protocol prevents the curator from observing future tasks when making a curation decision.

Holding the executor and task sequence fixed within a group makes group-relative reward comparisons conditional on the same execution setting, while varying the executor across groups exposes the curator to different behavioral patterns and skill needs. Multi-executor training is therefore not merely a way to increase data diversity: its purpose is to teach a shared curator how the mapping from execution experience to appropriate curation changes with executor behavior (Appendix~\ref{app:pool} discusses the remaining sources of variation).

\paragraph{Reward and group-relative advantage.}
We optimize the curator with GRPO \citep{shao2024deepseekmath}. Following the reward components of \citet{ouyang2026skillos}, rollout $n$ receives
\begin{equation}
R^{(n)}=r^{(n)}_{\mathrm{task}}+\lambda_f r^{(n)}_{\mathrm{func}}
+\lambda_c r^{(n)}_{\mathrm{comp}}+\lambda_u r^{(n)}_{\mathrm{qual}},
\qquad
A^{(n)}=\frac{R^{(n)}-\mu_G}{\sigma_G+\epsilon_A},
\label{eq:reward}
\end{equation}
where $\mu_G$ and $\sigma_G$ denote the mean and standard deviation across the $N$ rollouts in the same group, and $\epsilon_A>0$ stabilizes normalization. The task reward averages outcomes on $x_2,\ldots,x_T$ ($x_1$ is excluded because its repository is initially empty), and the auxiliary terms, averaged over curation steps, reward successfully executed curation calls, repository compactness, and content quality assessed by a training-time judge. We exclude descriptor tokens from the compression denominator so that a longer behavioral profile or model card does not itself increase the compression reward (Appendix~\ref{app:comp}).

\subsection{Retrieval- and Behavior-Aware Temporal Credit Assignment}
\label{sec:gates}

The rollout-level advantage evaluates the evolving repository as a whole, but does not identify which individual curation actions plausibly influenced later execution; applying it to every edit therefore introduces noisy supervision, especially when an edited skill is never retrieved again. To address this temporal credit assignment problem, EASE uses two approximate temporal-attribution gates to select curation actions with observable downstream influence (exact definitions and edge cases in Appendix~\ref{app:gates}).

\paragraph{Gate 1: retrieval incidence.}
For an edit at step $t$, let $Z_t$ count subsequent tasks for which a skill touched by $\Delta_t$ either enters retrieval or would have entered retrieval had the edit not been applied. To account for \texttt{delete} and \texttt{modify} operations, the test also considers a patched repository that restores the pre-edit skill state. Gate 1 admits an edit only if $Z_t>0$, excluding curation actions with no observed opportunity to influence later execution.

\paragraph{Gate 2: behavioral deviation.}
Retrieval alone does not establish that a skill affects executor behavior. For each retrieval-incident edit, we therefore compare subsequent skill-assisted action sequences against cached skill-free reference trajectories generated under the same executor and environment configuration, computing a normalized Levenshtein deviation while subtracting background variation among the skill-free runs. Let $\mathcal{D}_j$ denote this deviation on task $j$. Gate 2 admits an edit only if at least one later retrieval-incident task satisfies $\mathcal{D}_j>\tau$ for a threshold $\tau\ge0$. With both gates enabled, the admitted set for rollout $n$ is
\begin{equation}
\mathcal{M}^{(n)}=\left\{t: Z^{(n)}_t>0\ \text{and}\
\max_{j>t:\ \text{touched skill retrieved at }j}\mathcal{D}^{(n)}_j>\tau\right\};
\label{eq:admitted}
\end{equation}
variants that disable one or both gates are defined in Appendix~\ref{app:gates}. Because $\mathcal{D}_j$ measures whether skill-conditioned execution departs from the executor's skill-free behavior, rather than the marginal contribution of a particular edit, we interpret the two gates as approximate temporal attribution, rather than exact causal credit assignment.

\paragraph{Masked policy optimization.}
Let $q^{(n)}_{t,\ell}(\theta)$ denote the token-level importance ratio for token $\ell$ at curation step $t$, and let $A^{(n)}$ denote the rollout-level advantage. We maximize the clipped PPO surrogate only over admitted curation steps:
\begin{equation}
\mathcal{L}(\theta)=\frac{1}{|\mathcal{I}|}\sum_{n\in\mathcal{I}}
\frac{1}{|\mathcal{M}^{(n)}|}\sum_{t\in\mathcal{M}^{(n)}}
\frac{1}{L_t^{(n)}}\sum_{\ell=1}^{L_t^{(n)}}
\ell_{\mathrm{clip}}\!\left(q^{(n)}_{t,\ell}(\theta),A^{(n)}\right),
\label{eq:surrogate}
\end{equation}
where $\mathcal{I}=\{n:|\mathcal{M}^{(n)}|>0\}$ and $\ell_{\mathrm{clip}}$ is the standard PPO clipped surrogate. Advantages are computed from all $N$ rollouts before masking, so the rollout-level reward determines whether a trajectory is beneficial overall, while temporal attribution determines which curation actions receive policy-gradient supervision; Appendix~\ref{app:gates} details the handling of empty admitted sets and fixed masks.

\paragraph{Deployment-time adaptation.}
The quality judge, skill-free reference trajectories, and temporal-attribution gates are used only during training. At deployment, curator and executor parameters remain fixed, and EASE adapts solely through the behavioral profile $P_t(E)$ and the skill repository $\mathcal{S}_t$, which allows the same curator to serve executors that were not available during training.

\section{Experiments}
\label{sec:experiments}
We organize our experiments around four research questions (\textbf{RQ}s). \textbf{RQ1}: Can a single curator serve many executors, including ones it has never seen? \textbf{RQ2}: Does adapting online at deployment actually matter? \textbf{RQ3}: Does behavior-aware curation make skills more useful and cheaper to use? \textbf{RQ4}: Which components make our behavior-aware curation work?

\subsection{Experimental Settings}
\label{sec:experimental-setting}

\paragraph{Benchmarks.}
We evaluate on three text-based interactive benchmarks: ALFWorld~\citep{shridhar2021alfworld}, a
household environment with six task categories; ScienceWorld~\citep{wang2022scienceworld}, a
simulated laboratory whose tasks require multi-step scientific procedures such as measuring, heating,
and mixing; and WebShop~\citep{yao2022webshop}, a simulated e-commerce website where the agent searches
for and purchases products that match natural-language instructions.

\paragraph{Metric.}
On ALFWorld, we report success rate (SR) macro-averaged over task categories, so that each
category contributes equally regardless of its size (Eq.~\ref{eq:macro-sr}, Appendix~\ref{app:implement}). On ScienceWorld and WebShop, we report the standard benchmark score.

\paragraph{Executors.}
The curator is trained with three frozen executors,
$\mathcal{E}_{\mathrm{train}}=\{$Qwen3-8B, Qwen3-32B, GPT-OSS-120B$\}$
\citep{yang2025qwen3,openai2025gptoss}. To test transfer, we evaluate with three API executors
from other model families, Kimi K2.6, DeepSeek V4 Flash, and Gemini 3.5 Flash, which are excluded
from both training and checkpoint selection.

\paragraph{Baselines.}
\emph{Training-free} baselines keep all model weights fixed: No Memory runs the executor without any
memory or skills; AWM~\citep{wang2025awm}, ReasoningBank~\citep{ouyang2026reasoningbank},
MemP~\citep{fang2026memp}, and Memento-Skills~\citep{zhou2026mementoskills} augment the executor with
workflows, reasoning memories, procedural memories, or skills accumulated from experience.
\emph{Finetuned curators}: SkillOS~\citep{ouyang2026skillos} trains a skill curator with RL for a single
executor; we denote the three seen executors by $\alpha$ (Qwen3-8B), $\beta$ (Qwen3-32B), and
$\gamma$ (GPT-OSS-120B), so that SkillOS-$\alpha$ is the curator trained with Qwen3-8B, and so on.
SkillOS-base denotes the same curator without RL training. For unseen executors, we run all three transferred curators and report, for each executor and
benchmark, the highest score among them (SkillOS-\textit{best}).

\paragraph{Implementation.}
The curator is Qwen3-8B finetuned with LoRA~\citep{hu2022lora} and GRPO~\citep{shao2024deepseekmath};
at evaluation, all weights are frozen and each stream starts from an empty repository and profile
(Appendix~\ref{app:implement}).

\subsection{Transferability across Executors (RQ1)}
\label{sec:rq1}
\label{sec:main-results}

\begin{table}[t]
\centering
\caption{Main results on ALFWorld (ALF), ScienceWorld (SciW), and WebShop (WS). We report category-macro success rate (\%) on ALF and the benchmark score on SciW and WS; Avg.\ is their arithmetic mean. \textit{Upper}: the three executors in the curator's training pool ($\alpha$, $\beta$, $\gamma$). \textit{Lower}: three unseen executors; SkillOS-\textit{best} reports, per executor and benchmark, the highest score among SkillOS-$\alpha$, SkillOS-$\beta$, and SkillOS-$\gamma$. \textbf{Bold} and \underline{underline} mark the best and second-best results.}
\label{tab:main-agentic-results}
\vspace{9pt}
\small
\setlength{\tabcolsep}{4.2pt}
\renewcommand{\arraystretch}{1.15}
\resizebox{\linewidth}{!}{%
\begin{tabular}{l*{12}{c}}
\toprule
\multirow{2}{*}[-0.5ex]{\textbf{Method}}
  & \multicolumn{4}{c}{\textsc{Qwen3-8B} ($\alpha$)}
  & \multicolumn{4}{c}{\textsc{Qwen3-32B} ($\beta$)}
  & \multicolumn{4}{c}{\textsc{GPT-OSS-120B} ($\gamma$)} \\
\cmidrule(lr){2-5} \cmidrule(lr){6-9} \cmidrule(lr){10-13}
  & ALF & SciW & WS & Avg.
  & ALF & SciW & WS & Avg.
  & ALF & SciW & WS & Avg. \\
\midrule
\noalign{\vskip -\belowrulesep}
\multicolumn{13}{c}{\cellcolor[HTML]{F3F3F3}\rule[-0.8ex]{0pt}{3.15ex}\textit{Training-Free}} \\
\noalign{\vskip -\aboverulesep}
\midrule
No Memory                       & 39.6 & 24.9 & 33.3 & 32.6 & 46.3 & 32.1 & 41.5 & 40.0 & 51.9 & 35.1 & 43.6 & 43.5 \\
AWM                             & 41.0 & 24.3 & 30.7 & 32.0 & 47.1 & 33.5 & 41.9 & 40.8 & 50.1 & 36.4 & 44.8 & 43.8 \\
ReasoningBank                   & \underline{43.3} & \underline{25.7} & \underline{35.4} & \underline{34.8} & \underline{49.8} & 35.1 & 43.4 & 42.8 & \underline{53.7} & \textbf{38.1} & \textbf{46.8} & \textbf{46.2} \\
MemP                            & 42.7 & \textbf{26.4} & \textbf{35.7} & \textbf{34.9} & 48.2 & \textbf{36.8} & \textbf{45.3} & \underline{43.4} & \textbf{54.1} & 37.4 & \underline{45.3} & \underline{45.6} \\
Memento-Skills                  & \textbf{44.9} & 25.1 & 34.3 & \underline{34.8} & \textbf{50.1} & \underline{36.4} & \underline{44.8} & \textbf{43.8} & 53.2 & \underline{37.6} & 44.7 & 45.2 \\
\midrule
\noalign{\vskip -\belowrulesep}
\multicolumn{13}{c}{\cellcolor[HTML]{F3F3F3}\rule[-0.8ex]{0pt}{3.15ex}\textit{Finetuned Curators}} \\
\noalign{\vskip -\aboverulesep}
\midrule
SkillOS-$\alpha$         & \underline{50.0} & \underline{30.6} & \underline{40.6} & \underline{40.4} & 57.8 & 33.1 & 44.7 & 45.2 & 59.6 & 37.7 & 44.5 & 47.3 \\
SkillOS-$\beta$        & 45.0 & 25.3 & 36.4 & 35.6 & \underline{66.3} & \underline{39.2} & \textbf{49.3} & \underline{51.6} & 63.7 & 40.6 & 47.4 & 50.6 \\
SkillOS-$\gamma$       & 41.1 & 22.1 & 32.1 & 31.8 & 61.2 & 36.9 & 45.1 & 47.7 & \underline{69.6} & \underline{43.3} & \textbf{51.2} & \underline{54.7} \\
\rowcolor{oursrow}\textbf{EASE} (\textit{ours})   & \textbf{55.0} & \textbf{31.5} & \textbf{42.8} & \textbf{43.1} & \textbf{69.7} & \textbf{42.1} & \underline{48.9} & \textbf{53.6} & \textbf{73.1} & \textbf{46.4} & \underline{50.3} & \textbf{56.6} \\
\midrule\midrule
\multirow{2}{*}[-0.5ex]{\textbf{Method}}
  & \multicolumn{4}{c}{\textsc{Kimi K2.6}}
  & \multicolumn{4}{c}{\textsc{DeepSeek V4 Flash}}
  & \multicolumn{4}{c}{\textsc{Gemini 3.5 Flash}} \\
\cmidrule(lr){2-5} \cmidrule(lr){6-9} \cmidrule(lr){10-13}
  & ALF & SciW & WS & Avg.
  & ALF & SciW & WS & Avg.
  & ALF & SciW & WS & Avg. \\
\midrule
\noalign{\vskip -\belowrulesep}
\multicolumn{13}{c}{\cellcolor[HTML]{F3F3F3}\rule[-0.8ex]{0pt}{3.15ex}\textit{Training-Free}} \\
\noalign{\vskip -\aboverulesep}
\midrule
No Memory                       & 72.1 & 63.2 & 45.3 & 60.2 & 76.3 & 70.6 & 42.5 & 63.1 & 78.2 & 72.2 & 49.6 & 66.7 \\
AWM                             & 71.2 & 63.8 & 46.1 & 60.4 & 75.5 & 72.4 & 44.8 & 64.2 & 77.9 & 74.7 & 50.8 & 67.8 \\
ReasoningBank                   & \underline{75.1} & \underline{65.2} & \underline{48.1} & \underline{62.8} & 78.3 & \underline{74.0} & 45.9 & 66.1 & \underline{80.1} & 75.3 & 53.2 & \underline{69.5} \\
MemP                            & \textbf{76.3} & 64.1 & \textbf{48.5} & \textbf{63.0} & \underline{78.8} & 73.9 & \underline{46.1} & \underline{66.3} & 79.2 & \textbf{77.1} & \textbf{55.6} & \textbf{70.6} \\
Memento-Skills                  & 74.9 & \textbf{65.7} & 47.9 & \underline{62.8} & \textbf{79.4} & \textbf{74.5} & \textbf{46.3} & \textbf{66.7} & \textbf{80.7} & \underline{76.2} & \underline{54.9} & \textbf{70.6} \\
\midrule
\noalign{\vskip -\belowrulesep}
\multicolumn{13}{c}{\cellcolor[HTML]{F3F3F3}\rule[-0.8ex]{0pt}{3.15ex}\textit{Finetuned Curators}} \\
\noalign{\vskip -\aboverulesep}
\midrule
SkillOS-\textit{best}       & 75.3 & 70.2 & 51.3 & 65.6 & 82.1 & 76.3 & 49.1 & 69.2 & 85.3 & \textbf{79.1} & 54.7 & 73.0 \\
\rowcolor{oursrow}\textbf{EASE} (\textit{ours})   & \textbf{81.1} & \textbf{72.6} & \textbf{54.6} & \textbf{69.4} & \textbf{85.3} & \textbf{78.4} & \textbf{51.9} & \textbf{71.9} & \textbf{88.2} & 78.6 & \textbf{58.5} & \textbf{75.1} \\
\bottomrule
\end{tabular}}
\end{table}

\paragraph{Executor-specific curators do not transfer well.}
Among the three SkillOS curators, the one trained with a given executor performs best with that
executor on every benchmark (Table~\ref{tab:main-agentic-results}, upper). Mismatched pairings trail
the matched curator by 2.3--10.0 pp. On ALFWorld, for example, SkillOS-$\gamma$ reaches
41.1\% with Qwen3-8B, compared with 50.0\% for SkillOS-$\alpha$, whereas SkillOS-$\alpha$
reaches 59.6\% with GPT-OSS-120B, compared with 69.6\% for SkillOS-$\gamma$ (see also
Figure~\ref{fig:heatmap}). These gaps motivate conditioning curation decisions on the executor that
consumes the skills.

\paragraph{Seen executors.}
A single EASE curator obtains the best average score on all three training executors, improving on the matched specialist by 2.7, 2.0, and 1.9 pp, respectively. It is best on ALFWorld (+5.0, +3.4, and +3.5 pp) and ScienceWorld (+0.9, +2.9, and +3.1 pp) for all three executors. On WebShop, it is best with Qwen3-8B, but trails the matched specialist by 0.4 and 0.9 pp with Qwen3-32B and GPT-OSS-120B, while still outperforming all training-free baselines. EASE thus matches or exceeds executor-specific curators with a single curator.

\paragraph{Unseen executors.}
Without further training, EASE also obtains the best average score on all three held-out executors,
exceeding the strongest baseline, SkillOS-\textit{best}, by 3.8, 2.7, and 2.1 pp on Kimi K2.6,
DeepSeek V4 Flash, and Gemini 3.5 Flash, respectively (Table~\ref{tab:main-agentic-results}, lower).
It is best in eight of the nine benchmark--executor combinations; the exception is ScienceWorld with
Gemini 3.5 Flash, where it trails SkillOS-\textit{best} by 0.5 pp. On ALFWorld, its margins over the
strongest baseline are 4.8, 3.2, and 2.9 pp. Since SkillOS-\textit{best} is selected separately for every executor and benchmark, these
margins are measured against an optimistically selected reference. These results support transfer beyond the training pool.

\subsection{Online Adaptation at Deployment (RQ2)}
\label{sec:rq2}
\label{sec:profile-dynamics}

After a 64-task warm-up, continuing to update the profile improves ALFWorld SR over freezing it by
$3.8$, $2.2$, and $1.1\,pp$ for Qwen3-8B, Qwen3-32B, and GPT-OSS-120B, respectively, with skill curation
enabled in both conditions. Continued online curation with profile updates also outperforms a frozen
repository by $5.9$, $3.7$, and $2.4\,pp$. A 16-task profile window outperforms both a 4-task window and
full-history aggregation, while 16- and 64-task windows perform similarly (Appendix~\ref{app:profile-dynamics},
Table~\ref{tab:profile-dynamics}). Across six task orderings of the same task set, SR varies by at most
$3.3\,pp$ (Appendix~\ref{app:task-order}). These results support keeping both behavioral evidence and
skills current during deployment.

\subsection{Skill Utility and Efficiency (RQ3)}
\label{sec:rq3}
\label{sec:pending-measurements}

\begin{figure}[t]
\centering
\includegraphics[width=\linewidth]{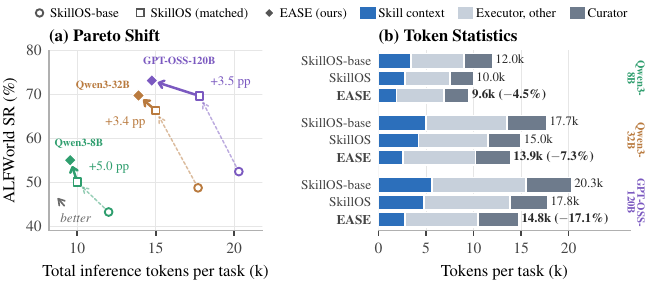}
\vspace{-2em}
\caption{\textbf{Higher success with fewer inference tokens on ALFWorld.}
(a) \textbf{\textit{Pareto shift}}: SR versus total inference tokens per task. (b) \textit{\textbf{Token statistics}}: per-task token composition; percentages give the change in total tokens
relative to the matched specialist.}
\label{fig:inference-efficiency}
\end{figure}

\paragraph{Performance and inference cost.}
Figure~\ref{fig:inference-efficiency}a shows that, on ALFWorld, EASE improves SR by $3.4\sim5.0\,pp$ over the
matched SkillOS specialists while using $4.5\sim17.1$\% fewer total deployment-time inference tokens
($10.7$\% fewer in aggregate over the three executors). The savings come mainly from shorter skill context
($32.1\sim41.7\%$ fewer tokens) and fewer environment steps, which more than offset the $150\sim250$ extra curator tokens spent on the descriptor (Figure~\ref{fig:inference-efficiency}b; Appendix~\ref{app:inference-cost}).

\paragraph{Retrieval and functional benefit.}
Averaged equally over Qwen3-8B, Qwen3-32B, and GPT-OSS-120B on ALFWorld, EASE reduces mean
repository size from 57.7 to 34.0 skills (41.0\% fewer) and raises the retrieval rate of edited skill
versions from $48.6\%$ to $66.7\%$ ($37.3\%$ relative) compared with the matched SkillOS specialists.
Paired edit-rollback diagnostics show a mean local SR benefit of 2.67 pp for EASE edits versus
1.67 pp for the specialists ($60.0\%$ relative). The same trends, including lower token usage, hold on
ScienceWorld and WebShop. Full measurements, definitions, and
Figure~\ref{fig:skill-utility} appear in Appendices~\ref{app:inference-cost} and~\ref{app:skill-utility}.

\subsection{Ablation Study (RQ4)}
\label{sec:rq4}
\label{sec:component-analysis}

\begin{figure}[t]
\centering
\includegraphics[width=\linewidth]{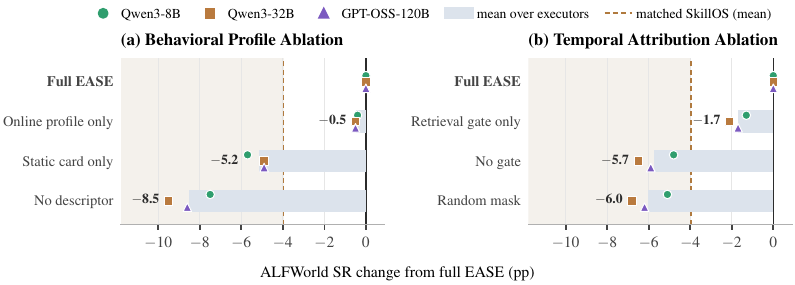}
\vspace{-2em}
\caption{\textbf{Component ablations on ALFWorld.} Markers show the perf. drop of each ablation variant from full EASE (Table~\ref{tab:main-agentic-results}); The dashed line marks the mean change of the matched SkillOS specialists relative to EASE ($-4.0\,pp$). 
}
\label{fig:component-ablations}
\end{figure}

All variants in Figure~\ref{fig:component-ablations} use the same executor pool and training budget.

\paragraph{Behavioral profile ablation.}
Removing the descriptor causes the largest drop, $7.5\sim9.5\,pp$, and places the pooled curator below
every matched SkillOS specialist. A static model card recovers only $1.8\sim4.6\,pp$ of this loss and still
trails the specialists, whereas the online profile alone comes within $0.5\,pp$ of the full descriptor
(Figure~\ref{fig:component-ablations}a; Table~\ref{tab:conditioning-analysis}). The benefit of conditioning
thus comes mainly from observed behavior rather than from executor identity.

\paragraph{Temporal attribution ablation.}
Removing both gates costs $4.8\sim6.5\, pp$. The retrieval gate recovers most
of this loss ($3.5\sim4.4 pp$), and the behavioral gate adds a further $1.3\sim2.1\,pp$. A random mask with the same admission rate performs on par with ungated training, so the gain comes from \emph{which} curation steps are updated (Figure~\ref{fig:component-ablations}b). 

\section{Conclusion}
We presented EASE, a framework for behavior-adaptive skill curation.
Unlike previous methods, EASE explicitly conditions on each executor's online behavior to curate customized skills at test time. Our experiments, across three benchmarks and six executors, have shown that EASE achieves the best average performance, improving on the strongest existing baseline by $1.9\sim3.8\,pp$, \textit{without} per-executor finetuning. Beyond performance, EASE curates more compact skill repository whose edited skills are retrieved $36.3\sim38.7$\% more often, while reducing total inference tokens by $9.1\sim14.5$\% across all three benchmarks.




\bibliography{iclr2027_conference}
\bibliographystyle{iclr2027_conference}

\appendix
\section{Extra Implementation Details}
\label{app:implement}

The implementation follows the online loop in Section~\ref{sec:method}: the executor
completes a task before the profile and repository are updated, and the resulting skills
are available only to later tasks.

\paragraph{Hyperparameters}

During RL finetuning, each GRPO update samples 32 groups, choosing
one executor per group with probability $1/3$ and collecting eight rollouts over the same
ordered stream of ten related training tasks. We retrieve at most $k=5$ skills and use a
profile window of $W=16$ unless varied in the profile diagnostics. The paragraphs below describe
the stream construction, reward settings, and optimization hyperparameters.
At evaluation, the curator's weights are frozen and skills
are edited online under the streaming protocol in \S\ref{sec:prelim}. Kimi K2.6,
DeepSeek V4 Flash, and Gemini 3.5 Flash are used only for transfer evaluation and are excluded
from both training and checkpoint selection. 
Tables~\ref{tab:method-hyperparameters}
and~\ref{tab:training-hyperparameters} collect the settings used for EASE.

\paragraph{Constructing experience for skill reuse.}
We use ALFWorld's six task categories to identify related training tasks. Each training
stream contains ten distinct tasks from one category, presented in random order; tasks
may be reused across streams. For each group, we sample Qwen3-8B, Qwen3-32B, or
GPT-OSS-120B with equal probability and keep that executor and task sequence fixed across
the eight rollouts. Each rollout starts with an empty repository and profile, so its
editing decisions depend only on its own execution history. Since the training stream
length $T=10$ is shorter than the profile window $W=16$, training profiles aggregate all
completed tasks in the current rollout. During longer evaluation streams, the window
retains only the most recent 16 completed tasks.

\begin{table}[htbp]
\centering
\small
\renewcommand{\arraystretch}{1.10}
\caption{Method and evaluation settings. The profile window is varied only in the
profile-dynamics diagnostics. ALFWorld evaluation tasks are drawn from its
134-task \texttt{valid unseen} split.}
\label{tab:method-hyperparameters}
\begin{tabular}{lr}
\toprule
\textbf{Setting} & \textbf{Value} \\
\midrule
Retrieved skills, $k$ & 5 \\
Profile window, $W$ & 16 \\
Rollouts per group, $N$ & 8 \\
Training tasks per stream, $T$ & 10 \\
Skill-free references per task and executor, $m$ & 4 \\
Behavioral-deviation threshold, $\tau$ & 0.05 \\
Function-call reward weight, $\lambda_f$ & 1.0 \\
Compression reward weight, $\lambda_c$ & 0.05 \\
Quality reward weight, $\lambda_u$ & 0.10 \\
\midrule
Environment actions per episode (maximum) & 50 \\
Task-order permutations & 6 \\
Profile-diagnostic warm-up, $t_0$ & 64 \\
Profile-diagnostic suffix length & 70 \\
Retrieval follow-up window, $H$ & 20 \\
Edit-rollback paired comparisons per executor & 200 \\
\bottomrule
\end{tabular}
\end{table}

\paragraph{Learning the transferable curator.}
We optimize LoRA parameters of Qwen3-8B with AdamW, keeping the underlying model and
all executors frozen. The adapters have rank 16, scaling factor 32, and
dropout 0.05, and are applied to the attention and feed-forward projections listed in
Table~\ref{tab:training-hyperparameters}.
Each sampling batch contains $B=32$ executor--stream groups,
each with $N=8$ rollouts; $B$ therefore counts groups before rollout expansion.
We perform one optimization epoch per rollout batch for 60 updates. The objective is
the masked surrogate in Eq.~(\ref{eq:surrogate-app}), without a KL penalty. Advantages are
normalized within each group using $\epsilon_A=10^{-6}$, before the gates select editing
steps for optimization.

\begin{table}[htbp]
\centering
\small
\renewcommand{\arraystretch}{1.10}
\caption{Curator optimization and quality-judge settings. Executor models and the judge
remain frozen throughout training; only the curator's LoRA parameters are optimized.}
\label{tab:training-hyperparameters}
\begin{tabular}{lr}
\toprule
\textbf{Setting} & \textbf{Value} \\
\midrule
Curator backbone & Qwen3-8B \\
Finetuning method & LoRA \\
LoRA rank & 16 \\
LoRA scaling factor & 32 \\
LoRA dropout & 0.05 \\
LoRA target modules & \begin{tabular}[t]{@{}r@{}}
\texttt{q\_proj}, \texttt{k\_proj}, \texttt{v\_proj},\\
\texttt{o\_proj}, \texttt{gate\_proj},\\
\texttt{up\_proj}, \texttt{down\_proj}
\end{tabular} \\
Optimizer & AdamW \\
Learning rate & $1\times10^{-6}$ \\
Groups per sampling batch, $B$ & 32 \\
Clipping parameter, $\varepsilon$ & 0.2 \\
Advantage normalization constant, $\epsilon_A$ & $1\times10^{-6}$ \\
KL coefficient, $\beta$ & 0 \\
Curator training temperature & 1.0 \\
Maximum input length (tokens) & 16,384 \\
Maximum output length (tokens) & 4,096 \\
Training updates & 60 \\
Optimization epochs per rollout batch & 1 \\
\midrule
Quality-judge backbone & Qwen3-32B \\
Judge temperature & 0 \\
Judge evaluations per editing step & 1 \\
Judge score range & $[0,1]$ \\
\bottomrule
\end{tabular}
\end{table}

\paragraph{Supervising edits through outcomes and quality.}
The task reward averages ALFWorld's binary environment-success signals on
$x_2,\ldots,x_T$. A frozen Qwen3-32B judge evaluates each editing step once at temperature
zero, assessing correctness, executability, and reusability on a $[0,1]$ scale. These
quality scores are averaged over editing steps and combined with the other reward terms
using the weights in Table~\ref{tab:method-hyperparameters}. For the behavioral gate,
we cache four skill-free reference trajectories per training task and executor, using
the same initial state and execution configuration as the skill-assisted runs, and set
$\tau=0.05$. Reference trajectories and judge scores provide training supervision only.

\paragraph{Following adaptation at evaluation.}
We use BM25 as retriever throughout this paper over skill keywords \citep{robertson2009bm25},
although EASE is not tied to a particular retriever.
Each executor is evaluated on a stream of all 134 \texttt{valid unseen} tasks, with an
empty repository and profile at the start. The task-order analysis uses six permutations
of a 120-task subset of this split (Appendix~\ref{app:task-order}).
ALFWorld success rate is macro-averaged over task categories:
\begin{equation}
    \mathrm{SR}_{\text{macro}} = \frac{1}{|\mathcal{C}|}\sum_{c \in \mathcal{C}}
    \frac{1}{|\mathcal{D}_c|}\sum_{i \in \mathcal{D}_c}\mathbf{1}[\text{task } i \text{ succeeds}],
    \label{eq:macro-sr}
\end{equation}
where $\mathcal{C}$ is the set of categories and $\mathcal{D}_c$ is the set of tasks in category $c$. For the profile diagnostics,
conditions branch from a common
history after task 64 and are scored on the remaining tasks, isolating continued
adaptation from the shared warm-up. Skill retrieval is measured over the next $H=20$
tasks for versions with a complete follow-up window. The edit-rollback diagnostics use paired 
comparisons per executor on tasks from the same evaluation split; these additional executions 
do not update the main evaluation history (Appendix~\ref{app:skill-utility}).
The ScienceWorld and WebShop efficiency and skill-utility measurements follow the same protocol
on a 100-task evaluation stream per executor. Edit-rollback diagnostics sample 50 edits per method and
executor, each paired on four tasks, and differences are measured in benchmark score points.

\section{Executor behavioral profile features}
\label{app:profile}

Table~\ref{tab:profile} lists the feature families of $P_t(E)$
(\S\ref{sec:profile}). Features use stored logs and environment signals without additional
model calls. Their availability depends on the environment: an action rejection may be directly
observed, whereas a semantic error type or implicit reliance on a skill may not be. Unavailable
features are marked as missing. Self-assessed correctness is recorded separately from
externally observed outcomes.

For the window in Eq.~(\ref{eq:profile-update}), rates pool event counts and their corresponding
exposure counts, rather than averaging episode-level percentages with unequal denominators.
A feature with no eligible observations is unavailable, not zero. Averages carry their sample
counts, and the profile records the number of completed episodes in its window. Updating the
profile after $x_t$ ensures that the edit $\Delta_t$ uses observations through $t$ only.

\begin{table}[htbp]
\caption{Executor behavioral profile features. Examples are included only when supported by observable
signals; retrieval alone is not evidence that a skill was followed.}
\label{tab:profile}
\begin{center}
\small
\begin{tabular}{lp{0.72\linewidth}}
\toprule
\textbf{Family} & \textbf{Example statistics and observability} \\
\midrule
Action validity & Rejected-action rate and parse failures from action validation. \\
Failure recovery & Repeated rejected actions; steps to the next accepted action, with
unrecovered episodes counted separately. \\
State tracking & Repeated visits to logged states; precondition violations when exposed by
the environment. \\
Efficiency & Steps per episode and frequency of reaching the interaction budget. \\
Error types & Subtask failures or reward-component losses when the environment exposes them. \\
Skill utilization & Retrieval frequency; explicit skill references or invocations when logged;
outcomes conditional on retrieval or observed use, reported separately. \\
\bottomrule
\end{tabular}
\end{center}
\end{table}

\section{Executor composition of GRPO groups}
\label{app:pool}

Within each GRPO group, all $N$ rollouts share the same frozen executor and ordered task stream.
Their repositories, profiles, and sampled execution histories evolve independently. The mean
in Eq.~(\ref{eq:reward}) therefore compares curation policies under a common executor and task
sequence, avoiding direct comparisons between the base success rates of different executors
within the same advantage calculation. It does not remove executor sampling noise, environment
stochasticity, or differences between repository histories.

The executor is sampled across groups to provide experience from multiple executors. Since even
one executor's profile changes over time, diversity is an additional source of variation in
conditioning, not a prerequisite for the descriptor to vary. Nor does diversity force the
curator to use the profile; removing or shuffling it provides an empirical test. Executors near
a benchmark's floor or ceiling may offer little variation in task outcomes, but the resulting
gradient magnitude also depends on auxiliary rewards and the normalization by
$\sigma_G+\epsilon_A$. Identical total rewards within a group yield zero advantages.

\section{Credit Assignment}
\label{app:gates}

\paragraph{Gate 1: retrieval incidence.}
For $j>t$, let $\mathcal{T}_{t,j}$ contain the before- and after-edit versions of skills changed by $\Delta_t$ that have not been edited again before $x_j$. Let $\Pi_{t,j}$ restore their pre-edit state in the current repository $\mathcal{S}_j$, leaving other skills unchanged. We define
\begin{equation}
\begin{aligned}
\rho^{(n)}_{t,j}&=\mathbf{1}\!\left[
\mathcal{T}^{(n)}_{t,j}\cap\left(
\mathcal{R}_k(x_j;\mathcal{S}^{(n)}_j)\cup
\mathcal{R}_k(x_j;\Pi_{t,j}(\mathcal{S}^{(n)}_j))\right)\neq\varnothing
\right],\\
Z^{(n)}_t&=\sum_{j=t+1}^{T}\rho^{(n)}_{t,j}.
\end{aligned}
\label{eq:rho}
\end{equation}
The patched retrieval permits credit for deleting a skill that would otherwise have been retrieved. Tracking versions avoids silently attributing a later rewrite to an earlier edit; it also omits effects inherited through subsequent edits. Appendix~\ref{app:incidence} specifies the patch and operation-wise tests. Gate 1 admits steps with $Z^{(n)}_t>0$.

\paragraph{Gate 2: behavioral deviation.}
Retrieval alone does not establish that a skill changes execution. For each training task and executor, we cache $m\geq2$ skill-free action sequences
$\{\tilde a_j^{(r)}\}_{r=1}^{m}$, using the same initial environment state, prompting, and decoding configuration as skill-assisted execution. We measure deviation from these references as
\begin{equation}
\mathcal{D}^{(n)}_j=\left[
\frac{1}{m}\sum_{r=1}^{m}d(a^{(n)}_j,\tilde a_j^{(r)})
-\binom{m}{2}^{-1}\sum_{r<r'}d(\tilde a_j^{(r)},\tilde a_j^{(r')})
\right]_+,
\label{eq:gain}
\end{equation}
where $[z]_+=\max(z,0)$ and $d$ is the Levenshtein distance between action sequences divided by
$\max(|a|,|b|,1)$. Each parsed action is one sequence element. The second term estimates variation among skill-free runs. Moreover, $\mathcal{D}_j$ measures deviation under the entire retrieved set, not improvement or the marginal effect of $\Delta_t$.

\paragraph{Combined admission.}
With both gates enabled, the admitted set is
\begin{equation}
\mathcal{M}^{(n)}=\left\{t: Z^{(n)}_t>0\ \text{and}\
\max_{j>t:\rho^{(n)}_{t,j}=1}\mathcal{D}^{(n)}_j>\tau\right\},\qquad \tau\geq0.
\label{eq:admitted-app}
\end{equation}
The maximum is evaluated only when $Z_t^{(n)}>0$. Disabling Gate 2 explicitly removes the second condition; setting $\tau=0$ still excludes zero deviation. Disabling both gates admits all curation steps. Reference trajectories are reused while their executor and environment configuration remains unchanged, with their inference cost accounted for as training overhead.

\paragraph{Masked policy optimization.}
Let $y^{(n)}_{t,\ell}$ be token $\ell$ of the curator's generated response at step $t$, of length
$L_t^{(n)}$, and let $h^{(n)}_{t,\ell}$ be its logged context, including preceding output and
function-call feedback. The token importance ratio is
$q^{(n)}_{t,\ell}(\theta)=\pi_{\theta}(y^{(n)}_{t,\ell}\mid h^{(n)}_{t,\ell})/
\pi_{\theta_{\mathrm{old}}}(y^{(n)}_{t,\ell}\mid h^{(n)}_{t,\ell})$.
Using the clipped surrogate of proximal policy optimization \citep{schulman2017ppo},
we maximize the following masked objective over
$\mathcal{I}=\{n:|\mathcal{M}^{(n)}|>0\}$:
\begin{equation}
\begin{aligned}
\mathcal{L}(\theta)&=\frac{1}{|\mathcal{I}|}\sum_{n\in\mathcal{I}}
\frac{1}{|\mathcal{M}^{(n)}|}\sum_{t\in\mathcal{M}^{(n)}}
\frac{1}{L_t^{(n)}}\sum_{\ell=1}^{L_t^{(n)}}
\ell_{\mathrm{clip}}\!\left(q^{(n)}_{t,\ell}(\theta),A^{(n)}\right),\\
\ell_{\mathrm{clip}}(q,A)&=\min\!\left(qA,
\operatorname{clip}(q,1-\varepsilon,1+\varepsilon)A\right).
\end{aligned}
\label{eq:surrogate-app}
\end{equation}
Empty admitted sets are omitted from the loss, and the update is skipped if $\mathcal{I}$ is
empty; advantages are always computed from the original $N$ rollouts. Masks are held fixed
during optimization. This applies the same composite advantage to every admitted step, so
masking also removes direct auxiliary-reward supervision for excluded steps, including the
final edit when no subsequent task is available. We retain this tradeoff explicitly rather
than treating the gates as exact credit assignment.

\section{Retrieval-incidence tests}
\label{app:incidence}

The patch in Eq.~(\ref{eq:rho}) is a local repository intervention, not a replay of the stream
without an earlier edit. We record skill identifiers and before- and after-edit versions for
each successfully applied operation. For a sequence of calls $\Delta_t$, the relevant change is
its net effect on each identifier; failed calls and sequences with no net repository change
contribute no touched versions.

For a later task $x_j$, an identifier remains eligible for the test only if no successful
operation has touched it between steps $t+1$ and $j-1$. Let $U_{t,j}$ be the eligible identifiers
whose state changed at $t$. The set $\mathcal{T}_{t,j}$ contains their nonempty pre- and
post-edit versions. The patch $\Pi_{t,j}$ replaces the current state of every identifier in
$U_{t,j}$ with its pre-edit state, leaving all other entries unchanged. Restoring an absent
pre-edit state removes an added skill; restoring a deleted skill inserts its saved version.
Version comparisons include the identifier, header, and body. This rule excludes superseded
edits from direct attribution, including cases where a later edit reinstates identical text.

For an isolated change that remains eligible at $j$, the tests reduce to
\begin{align*}
\texttt{add}(s):\quad &s\in\mathcal{R}_k(x_j;\mathcal{S}_j),\\
\texttt{delete}(s):\quad &s\in\mathcal{R}_k(x_j;\mathcal{S}_j\cup\{s\}),\\
\texttt{modify}(s\to s'):\quad &s'\in\mathcal{R}_k(x_j;\mathcal{S}_j)
\ \text{or}\ s\in\mathcal{R}_k(x_j;(\mathcal{S}_j\setminus\{s'\})\cup\{s\}).
\end{align*}
A step containing several changes uses the joint patch defined above. Both actual and patched
retrieval use the repository at $j$, rather than the stale repository at $t$, and use the same
retrieval settings and tie-breaking rule. Actual retrieval is already logged; evaluating a
nontrivial joint patch requires one additional retrieval query per $(t,j)$ pair. Across a
stream, the number of such queries can grow quadratically with its length. These checks add no
executor calls, but their retrieval and version-storage costs should still be measured.

\section{Compression reward under executor conditioning}
\label{app:comp}

Let $\chi_t$ be the curator input and let $\chi_t^{\mathrm{base}}$ be the same input with the
descriptor field $e_t$ removed. Writing $\operatorname{len}_{\mathrm{tok}}$ for token length
under a fixed tokenizer, the rollout-level compression term is
\begin{equation}
r_{\mathrm{comp}}^{(n)}=\frac{1}{T}\sum_{t=1}^{T}\left(
1-\frac{\operatorname{len}_{\mathrm{tok}}(\mathcal{S}_{t+1}^{(n)})}
{\max\{1,\operatorname{len}_{\mathrm{tok}}(\chi_t^{\mathrm{base},(n)})\}}
\right).
\label{eq:compression}
\end{equation}
The numerator measures the repository after applying $\Delta_t$; the denominator measures
the descriptor-free input used to produce it. Excluding both the profile and static metadata
prevents their length from directly increasing this reward. The term remains a relative
compression measure: differences in trajectories and retrieved context can still change its
denominator. It is neither a fixed token budget nor a direct measurement of retrieval cost,
which must be reported separately.

\section{Curator Conditioning Comparisons}
\label{app:conditioning-analysis}

\paragraph{Conditioning a skill curator.}
Table~\ref{tab:conditioning-analysis} summarizes the conditioned curator configurations.
These results complement the executor-specific references in
Table~\ref{tab:main-agentic-results}.

\begin{table}[htbp]
\centering
\small
\setlength{\tabcolsep}{6pt}
\renewcommand{\arraystretch}{1.15}
\caption{Conditioned curator comparisons on ALFWorld. Entries are category-macro SR
(\%). All configurations enable both temporal-attribution gates; the first two rows correspond to
the ``No descriptor'' and ``Static card only'' variants in Figure~\ref{fig:component-ablations}a.}
\label{tab:conditioning-analysis}
\begin{tabular}{lccc}
\toprule
\multirow{2}{*}{\textbf{Configuration}}
    & \multicolumn{3}{c}{\textbf{Executors}} \\
\cmidrule(lr){2-4}
    & \textsc{Qwen3-8B} & \textsc{Qwen3-32B} & \textsc{GPT-OSS-120B} \\
\midrule
Pooled, unconditioned & 47.5 & 60.2 & 64.5 \\
Static-card conditioning & 49.3 & 64.8 & 68.2 \\
EASE (\textit{ours}) & \textbf{55.0} & \textbf{69.7} & \textbf{73.1} \\
\bottomrule
\end{tabular}
\end{table}

\paragraph{Pooling helps, but leaves a gap to executor-specific training.}
The unconditioned pooled curator improves over SkillOS-base by 4.3, 11.5, and 12.1 pp on
Qwen3-8B, Qwen3-32B, and GPT-OSS-120B, respectively, but remains 2.5, 6.1, and 5.1 pp below the
corresponding specialists. Exposure to trajectories from multiple executors alone therefore does not
close the gap. Adding a static model card raises SR by a further 1.8, 4.6, and 3.7 pp, yet the
static-card variant still trails the matched specialists by 0.7, 1.5, and 1.4 pp. Identifying the
executor helps, but identity alone is not sufficient.

The complete EASE configuration improves on the unconditioned pooled curator by 7.5, 9.5, and 8.6 pp,
and on the static-card variant by 5.7, 4.9, and 4.9 pp. Since all three
configurations enable both temporal-attribution gates, the difference between the static-card
variant and the complete configuration isolates the contribution of the online profile, which is
what lifts the shared curator above the executor-specific specialists.

\section{Online Adaptation and Profile Dynamics}
\label{app:profile-dynamics}

\paragraph{Recent behavior and continued profile updates.}
Table~\ref{tab:profile-dynamics} examines deployment-time adaptation using the EASE curator
evaluated in Table~\ref{tab:main-agentic-results}. After a common warm-up ending at task $t_0=64$, each condition starts from a copy
of the same repository and interaction history and follows the same remaining task order.
Freezing the profile while continuing skill edits tests the value of fresh behavioral
evidence; freezing the repository tests the value of continued online editing. The remaining
conditions compare profile windows of 4, 16, and 64 tasks with full-history aggregation.
Curator and executor weights remain fixed throughout. Scores are category-macro SR on the
common 70-task suffix of the 134-task \texttt{valid unseen} stream. These scores are reported
separately from the full-stream main evaluation; with the reference configuration, SR on the
70-task suffix exceeds SR on the 64-task warm-up by 6.3, 5.4, and 4.4 pp on Qwen3-8B, Qwen3-32B, and
GPT-OSS-120B, consistent with the benefit of accumulated experience. Additional task-order
diagnostics are reported in Appendix~\ref{app:task-order}.

Updating the profile with $W=16$ exceeds freezing it at $t_0$ by 3.8, 2.2, and 1.1 pp,
respectively. Since both conditions continue editing skills, the comparison supports the
value of keeping behavioral evidence current as the repository evolves. Relative to a frozen
repository, the reference configuration gains 5.9, 3.7, and 2.4 pp. These are complementary
deployment interventions, rather than additive estimates of profile and repository effects.

The reference window also exceeds full-history aggregation by 1.3, 0.9, and 0.5 pp and the
four-task window by 1.7, 1.0, and 0.6 pp. This pattern is consistent with retaining enough
recent experience to characterize recurring behavior while limiting the influence of older
interactions. The 16- and 64-task windows are close: $W=64$ leads by 0.3 pp on Qwen3-8B
and 0.1 pp on GPT-OSS-120B, while $W=16$ leads by 0.2 pp on Qwen3-32B. The results support
continued profile updates and a finite recent history, without identifying one window as
uniformly best across executors.

\begin{table}[htbp]
\centering
\small
\setlength{\tabcolsep}{5pt}
\renewcommand{\arraystretch}{1.15}
\caption{Profile-dynamics diagnostics. Entries report category-macro SR (\%)
on the common 70-task suffix following a 64-task warm-up in the 134-task
\texttt{valid unseen} stream. Except for the frozen-repository control, all
conditions continue editing skills. The short, reference, and long windows contain 4, 16,
and 64 completed tasks, respectively.}
\label{tab:profile-dynamics}
\begin{tabular}{lccc}
\toprule
\multirow{2}{*}{\textbf{Configuration}}
    & \multicolumn{3}{c}{\textbf{Executors}} \\
\cmidrule(lr){2-4}
    & \textsc{Qwen3-8B} & \textsc{Qwen3-32B} & \textsc{GPT-OSS-120B} \\
\midrule
Updating profile, $W_{\mathrm{ref}} = 16$ & 58.0 & \textbf{72.3} & 75.2 \\
Profile frozen at $t_0$, $W_{\mathrm{ref}} = 16$ & 54.2 & 70.1 & 74.1 \\
Updating profile, $W_{\mathrm{short}} = 4$ & 56.3 & 71.3 & 74.6 \\
Updating profile, $W_{\mathrm{long}} = 64$ & \textbf{58.3} & 72.1 & \textbf{75.3} \\
Updating profile, full history & 56.7 & 71.4 & 74.7 \\
Repository frozen at $t_0$ & 52.1 & 68.6 & 72.8 \\
\bottomrule
\end{tabular}
\end{table}

\section{Additional Efficiency Measurements}
\label{app:inference-cost}

Table~\ref{tab:inference-cost} provides the absolute measurements underlying
Figure~\ref{fig:inference-efficiency}, together with skill-context and profile/card token counts
and environment steps. Token counts and steps are averages over all evaluation tasks,
including failures, with runtime settings held fixed within each executor comparison.
Curator and executor counts each include input and output tokens. Skill-context tokens are
part of executor input, and profile/card tokens are part of curator input; neither is added
again to the total. Environment steps measure execution effort. Reference-bank generation
and gate computation are training overhead and are excluded from these deployment measurements.

Relative to the matched specialists, EASE uses 150, 200, and 250 more curator tokens
per task, but 600, 1,300, and 3,300 fewer executor tokens on Qwen3-8B, Qwen3-32B, and
GPT-OSS-120B, respectively, reducing total tokens by 4.5\%, 7.3\%, and 17.1\% (10.7\% in
aggregate). Skill-context tokens decrease by 32.1\%, 38.1\%, and 41.7\%,
and environment steps decrease for all three executors. These counts do not measure latency,
monetary cost, or the additional cost of training.

Table~\ref{tab:inference-cost-sciworld} reports the same measurements on ScienceWorld. EASE
uses 250 more curator tokens per task than the matched specialists, but 4,000, 5,000, and 6,000
fewer executor tokens on Qwen3-8B, Qwen3-32B, and GPT-OSS-120B, respectively, reducing total
tokens by 13.4\%, 14.7\%, and 15.1\% (14.5\% in aggregate). Skill-context tokens decrease by
32.3\%, 33.3\%, and 34.1\%, and environment steps again decrease for all three executors.
Table~\ref{tab:inference-cost-webshop} reports the corresponding WebShop measurements. EASE again
uses 250 more curator tokens per task, but 3,500, 3,500, and 3,000 fewer executor tokens, reducing
total tokens by 10.8\%, 9.6\%, and 7.4\% (9.1\% in aggregate); skill-context tokens decrease by
30.8\%, 30.6\%, and 31.3\%, and environment steps decrease for all three executors. On WebShop,
these savings are not accompanied by a higher score on every executor: EASE trails the matched
specialist by 0.4 and 0.9 points with Qwen3-32B and GPT-OSS-120B
(Table~\ref{tab:main-agentic-results}).

\begin{table}[!htbp]
\centering
\small
\setlength{\tabcolsep}{6pt}
\renewcommand{\arraystretch}{1.10}
\caption{Performance and inference-token measurements on ALFWorld. SR for SkillOS and EASE matches
Table~\ref{tab:main-agentic-results}. Token counts and steps are per-task averages; skill-context and
descriptor tokens are included in the curator/executor totals. Dashes denote the absence of a
profile/card input.}
\label{tab:inference-cost}
\begin{tabular}{lrrr}
\toprule
\textbf{Metric} & \textbf{SkillOS-base} & \textbf{SkillOS} & \textbf{EASE} \\
\midrule
\multicolumn{4}{c}{\textit{Executor: \textsc{Qwen3-8B}}} \\
\midrule
SR (\%) & 43.2 & 50.0 & 55.0 \\
Curator tokens / task & 3k & 2.5k & 2.65k \\
Executor tokens / task & 9k & 7.5k & 6.9k \\
Skill-context tokens / task & 3.4k & 2.8k & 1.9k \\
Profile/card tokens / task & {-} & {-} & 245 \\
Environment steps / task & 28.5 & 25.0 & 22.0 \\
\midrule
\multicolumn{4}{c}{\textit{Executor: \textsc{Qwen3-32B}}} \\
\midrule
SR (\%) & 48.7 & 66.3 & 69.7 \\
Curator tokens / task & 4.2k & 3.5k & 3.7k \\
Executor tokens / task & 13.5k & 11.5k & 10.2k \\
Skill-context tokens / task & 5k & 4.2k & 2.6k \\
Profile/card tokens / task & {-} & {-} & 255 \\
Environment steps / task & 27.5 & 24.0 & 21.5 \\
\midrule
\multicolumn{4}{c}{\textit{Executor: \textsc{GPT-OSS-120B}}} \\
\midrule
SR (\%) & 52.4 & 69.6 & 73.1 \\
Curator tokens / task & 4.8k & 4k & 4.25k \\
Executor tokens / task & 15.5k & 13.8k & 10.5k \\
Skill-context tokens / task & 5.6k & 4.8k & 2.8k \\
Profile/card tokens / task & {-} & {-} & 260 \\
Environment steps / task & 26.5 & 23.0 & 20.5 \\
\bottomrule
\end{tabular}
\end{table}

\begin{table}[!htbp]
\centering
\small
\setlength{\tabcolsep}{6pt}
\renewcommand{\arraystretch}{1.10}
\caption{Performance and inference-token measurements on ScienceWorld. Scores for SkillOS and EASE
match Table~\ref{tab:main-agentic-results}. Token counts and steps are per-task averages; skill-context
and descriptor tokens are included in the curator/executor totals. Dashes denote the absence of a
profile/card input.}
\label{tab:inference-cost-sciworld}
\begin{tabular}{lrrr}
\toprule
\textbf{Metric} & \textbf{SkillOS-base} & \textbf{SkillOS} & \textbf{EASE} \\
\midrule
\multicolumn{4}{c}{\textit{Executor: \textsc{Qwen3-8B}}} \\
\midrule
Score & 25.8 & 30.6 & 31.5 \\
Curator tokens / task & 5.8k & 5k & 5.25k \\
Executor tokens / task & 27k & 23k & 19k \\
Skill-context tokens / task & 7.8k & 6.2k & 4.2k \\
Profile/card tokens / task & {-} & {-} & 250 \\
Environment steps / task & 32.5 & 28.7 & 24.4 \\
\midrule
\multicolumn{4}{c}{\textit{Executor: \textsc{Qwen3-32B}}} \\
\midrule
Score & 34.0 & 39.2 & 42.1 \\
Curator tokens / task & 6.2k & 5.4k & 5.65k \\
Executor tokens / task & 31k & 27k & 22k \\
Skill-context tokens / task & 9.2k & 7.5k & 5k \\
Profile/card tokens / task & {-} & {-} & 255 \\
Environment steps / task & 34.2 & 30.5 & 26.5 \\
\midrule
\multicolumn{4}{c}{\textit{Executor: \textsc{GPT-OSS-120B}}} \\
\midrule
Score & 37.0 & 43.3 & 46.4 \\
Curator tokens / task & 7k & 6k & 6.25k \\
Executor tokens / task & 37k & 32k & 26k \\
Skill-context tokens / task & 10.5k & 8.5k & 5.6k \\
Profile/card tokens / task & {-} & {-} & 260 \\
Environment steps / task & 36.0 & 31.8 & 27.3 \\
\bottomrule
\end{tabular}
\end{table}

\begin{table}[!htbp]
\centering
\small
\setlength{\tabcolsep}{6pt}
\renewcommand{\arraystretch}{1.10}
\caption{Performance and inference-token measurements on WebShop. Scores for SkillOS and EASE
match Table~\ref{tab:main-agentic-results}. Token counts and steps are per-task averages; skill-context
and descriptor tokens are included in the curator/executor totals. Dashes denote the absence of a
profile/card input.}
\label{tab:inference-cost-webshop}
\begin{tabular}{lrrr}
\toprule
\textbf{Metric} & \textbf{SkillOS-base} & \textbf{SkillOS} & \textbf{EASE} \\
\midrule
\multicolumn{4}{c}{\textit{Executor: \textsc{Qwen3-8B}}} \\
\midrule
Score & 35.2 & 40.6 & 42.8 \\
Curator tokens / task & 6k & 5.2k & 5.45k \\
Executor tokens / task & 28k & 25k & 21.5k \\
Skill-context tokens / task & 8k & 6.5k & 4.5k \\
Profile/card tokens / task & {-} & {-} & 250 \\
Environment steps / task & 18.5 & 16.0 & 13.5 \\
\midrule
\multicolumn{4}{c}{\textit{Executor: \textsc{Qwen3-32B}}} \\
\midrule
Score & 44.5 & 49.3 & 48.9 \\
Curator tokens / task & 6.5k & 5.7k & 5.95k \\
Executor tokens / task & 32k & 28k & 24.5k \\
Skill-context tokens / task & 9k & 7.2k & 5k \\
Profile/card tokens / task & {-} & {-} & 255 \\
Environment steps / task & 19.5 & 17.0 & 14.5 \\
\midrule
\multicolumn{4}{c}{\textit{Executor: \textsc{GPT-OSS-120B}}} \\
\midrule
Score & 46.5 & 51.2 & 50.3 \\
Curator tokens / task & 7k & 6.2k & 6.45k \\
Executor tokens / task & 36k & 31k & 28k \\
Skill-context tokens / task & 10k & 8k & 5.5k \\
Profile/card tokens / task & {-} & {-} & 260 \\
Environment steps / task & 21.0 & 18.5 & 16.0 \\
\bottomrule
\end{tabular}
\end{table}

\section{Skill Utility Measurements}
\label{app:skill-utility}

\begin{figure}[!htbp]
\centering
\includegraphics[width=\linewidth]{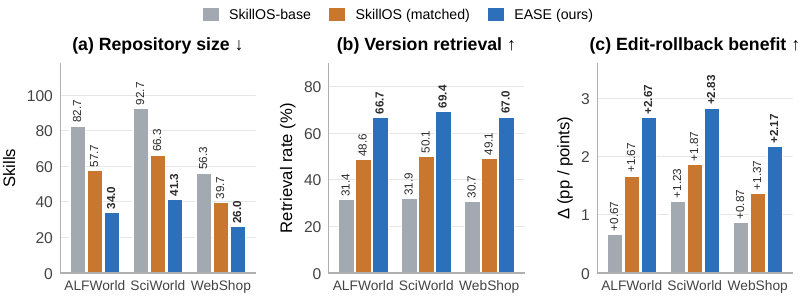}
\caption{\textbf{Compact repositories with higher retrieval and edit utility.}
Bars show equal-weight means across Qwen3-8B, Qwen3-32B, and GPT-OSS-120B on ALFWorld,
ScienceWorld, and WebShop. SkillOS denotes the executor-matched specialists. Each panel retains its original units;
the edit-rollback benefit is measured in SR percentage points on ALFWorld and in score points on
ScienceWorld and WebShop; arrows indicate the preferred direction. Complete per-executor measurements and metric
definitions are provided in Tables~\ref{tab:skill-utility}, \ref{tab:skill-utility-sciworld},
and~\ref{tab:skill-utility-webshop}.}
\label{fig:skill-utility}
\end{figure}

Table~\ref{tab:skill-utility} provides detailed executor-level measurements underlying
Figure~\ref{fig:skill-utility}. The figure takes the arithmetic mean of the three executor
entries for each method and metric. For fair comparison, the method ordering is the same 
within each executor for all three metrics. Tables~\ref{tab:skill-utility-sciworld} and~\ref{tab:skill-utility-webshop}
report the same measurements on ScienceWorld and WebShop, where the edit-rollback benefit is
measured in score points. Averaged over executors, EASE reduces repository size relative to the
matched specialists by 41.0\%, 37.7\%, and 34.5\% on ALFWorld, ScienceWorld, and WebShop,
raises the version retrieval rate by 37.3\%, 38.7\%, and 36.3\% (relative), and increases the
edit-rollback benefit by 60.0\%, 51.8\%, and 58.5\% (relative).

For each added or modified skill version with a complete $H=20$-task follow-up window,
we record whether that version is retrieved at least once during the window. Specifically,
retrieval of a later rewrite is not credited to an earlier version. 
Versions with fewer than 20 subsequent tasks available
in the evaluation stream are excluded from this metric for fair comparison.
Skill repository size is averaged over task boundaries before averaging across executors.

To assess functional benefit, we conduct paired comparisons per executor on hold-out tasks for evaluation. 
Each comparison pairs an edited repository
with the same repository after rolling back the selected edit. (Rollback removes an addition,
restores the previous version for a modification, or reinstates a deleted skill.) The mean
paired difference in task success, expressed in percentage points, measures the local benefit of the edit;
diagnostic trajectories are not fed back into the main evaluation stream.

These local differences support the functional value of the evaluated edits; they do not
decompose the full-stream performance gain or establish that every retrieved skill is
beneficial. Together with the direct context measurements in
Appendix~\ref{app:inference-cost}, Table~\ref{tab:inference-cost}, the results support compact
repositories whose edits are more often retrieved and yield larger measured local benefits.

\begin{table}[!htbp]
\centering
\small
\setlength{\tabcolsep}{6pt}
\renewcommand{\arraystretch}{1.15}
\caption{Skill utilization rate comparison on ALFWorld.}
\label{tab:skill-utility}
\begin{tabular}{lrrr}
\toprule
\textbf{Metric} & \textbf{SkillOS-base} & \textbf{SkillOS} & \textbf{EASE} \\
\midrule
\multicolumn{4}{c}{\textit{Executor: \textsc{Qwen3-8B}}} \\
\midrule
Version retrieval rate (\%) & 29.5 & 47.2 & 64.6 \\
Edit-rollback $\Delta$SR ($pp$) & +0.5 & +1.5 & +3.0 \\
Repository size (skills) & 85 & 60 & 35 \\
\midrule
\multicolumn{4}{c}{\textit{Executor: \textsc{Qwen3-32B}}} \\
\midrule
Version retrieval rate (\%) & 31.4 & 48.6 & 67.4 \\
Edit-rollback $\Delta$SR ($pp$) & +0.5 & +1.5 & +2.5 \\
Repository size (skills) & 83 & 58 & 34 \\
\midrule
\multicolumn{4}{c}{\textit{Executor: \textsc{GPT-OSS-120B}}} \\
\midrule
Version retrieval rate (\%) & 33.3 & 50.0 & 68.2 \\
Edit-rollback $\Delta$SR ($pp$) & +1.0 & +2.0 & +2.5 \\
Repository size (skills) & 80 & 55 & 33 \\
\bottomrule
\end{tabular}
\end{table}

\begin{table}[!htbp]
\centering
\small
\setlength{\tabcolsep}{6pt}
\renewcommand{\arraystretch}{1.15}
\caption{Skill utilization rate comparison on ScienceWorld. Retrieval rates are computed over
115/88/60, 108/82/56, and 104/78/54 complete-window versions (SkillOS-base/SkillOS/EASE) for
Qwen3-8B, Qwen3-32B, and GPT-OSS-120B, respectively; edit-rollback benefits use 50 sampled edits
per method and executor, each paired on four tasks.}
\label{tab:skill-utility-sciworld}
\begin{tabular}{lrrr}
\toprule
\textbf{Metric} & \textbf{SkillOS-base} & \textbf{SkillOS} & \textbf{EASE} \\
\midrule
\multicolumn{4}{c}{\textit{Executor: \textsc{Qwen3-8B}}} \\
\midrule
Version retrieval rate (\%) & 30.4 & 48.9 & 68.3 \\
Edit-rollback $\Delta$Score (points) & +0.6 & +1.1 & +1.8 \\
Repository size (skills) & 98 & 70 & 44 \\
\midrule
\multicolumn{4}{c}{\textit{Executor: \textsc{Qwen3-32B}}} \\
\midrule
Version retrieval rate (\%) & 31.5 & 50.0 & 69.6 \\
Edit-rollback $\Delta$Score (points) & +1.4 & +2.0 & +3.1 \\
Repository size (skills) & 92 & 66 & 41 \\
\midrule
\multicolumn{4}{c}{\textit{Executor: \textsc{GPT-OSS-120B}}} \\
\midrule
Version retrieval rate (\%) & 33.7 & 51.3 & 70.4 \\
Edit-rollback $\Delta$Score (points) & +1.7 & +2.5 & +3.6 \\
Repository size (skills) & 88 & 63 & 39 \\
\bottomrule
\end{tabular}
\end{table}

\begin{table}[!htbp]
\centering
\small
\setlength{\tabcolsep}{6pt}
\renewcommand{\arraystretch}{1.15}
\caption{Skill utilization rate comparison on WebShop. Retrieval rates are computed over
95/76/55, 91/75/52, and 88/71/47 complete-window versions (SkillOS-base/SkillOS/EASE) for
Qwen3-8B, Qwen3-32B, and GPT-OSS-120B, respectively; edit-rollback benefits use 50 sampled edits
per method and executor, each paired on four tasks.}
\label{tab:skill-utility-webshop}
\begin{tabular}{lrrr}
\toprule
\textbf{Metric} & \textbf{SkillOS-base} & \textbf{SkillOS} & \textbf{EASE} \\
\midrule
\multicolumn{4}{c}{\textit{Executor: \textsc{Qwen3-8B}}} \\
\midrule
Version retrieval rate (\%) & 30.5 & 47.4 & 65.5 \\
Edit-rollback $\Delta$Score (points) & +0.6 & +1.0 & +1.7 \\
Repository size (skills) & 58 & 41 & 27 \\
\midrule
\multicolumn{4}{c}{\textit{Executor: \textsc{Qwen3-32B}}} \\
\midrule
Version retrieval rate (\%) & 30.8 & 49.3 & 67.3 \\
Edit-rollback $\Delta$Score (points) & +0.9 & +1.4 & +2.2 \\
Repository size (skills) & 57 & 40 & 26 \\
\midrule
\multicolumn{4}{c}{\textit{Executor: \textsc{GPT-OSS-120B}}} \\
\midrule
Version retrieval rate (\%) & 30.7 & 50.7 & 68.1 \\
Edit-rollback $\Delta$Score (points) & +1.1 & +1.7 & +2.6 \\
Repository size (skills) & 54 & 38 & 25 \\
\bottomrule
\end{tabular}
\end{table}

\section{Task-Order Sensitivity}
\label{app:task-order}

\paragraph{Sensitivity to task order.}
To test whether EASE is sensitive to the order in which tasks arrive, we sample a fixed set $S$ of
120 tasks covering all six categories of the ALFWorld held-out split and construct six evaluation
streams that contain the same tasks in different orders, $P_t=\texttt{permute}_t(S)$ for
$t\in\{1,\ldots,6\}$. With the trained EASE curator and Qwen3-8B as the executor, each stream is
solved independently from an empty repository and profile, without prior knowledge of the ordering.
Table~\ref{tab:task-order} reports the SR of each stream.

\begin{table}[htbp]
\centering
\small
\renewcommand{\arraystretch}{1.15}
\caption{Task-order analysis over six permutations of the same 120 ALFWorld tasks with Qwen3-8B as
the executor. SR is the stream-level success rate (\%).}
\label{tab:task-order}
\begin{tabular}{lrrrrrr}
\toprule
\textbf{Task order} & 1 & 2 & 3 & 4 & 5 & 6 \\
\midrule
SR & 57.5 & 56.7 & 55.8 & 57.5 & 57.5 & 54.2 \\
\bottomrule
\end{tabular}
\end{table}

SR ranges from 54.2\% to 57.5\% (a spread of 3.3 pp) with no extreme outliers, suggesting that EASE
is not overly sensitive to the specific ordering of tasks. Real-world task streams, however, can mix
domains and difficulty levels in ways that are hard to model, and a more thorough study of ordering
effects is left to future work.

\end{document}